\documentclass[10pt,twocolumn,letterpaper]{article}

\usepackage[pagenumbers]{cvpr} 

\definecolor{cvprblue}{rgb}{0.21,0.49,0.74}
\usepackage[pagebackref,breaklinks,colorlinks,allcolors=cvprblue]{hyperref}
\usepackage{placeins}
\usepackage{tabularx}
\usepackage[utf8]{inputenc} 
\usepackage[T1]{fontenc}    
\usepackage{hyperref}       
\usepackage{url}            
\hypersetup{breaklinks,colorlinks,linkcolor=red,anchorcolor=blue,citecolor=cyan,filecolor=blue,urlcolor=blue}

\usepackage{booktabs}          
\usepackage{multirow}          
\usepackage{colortbl}          
\usepackage{arydshln}          
\usepackage{array}             

\usepackage{amsfonts}       
\usepackage{amsmath}	    
\usepackage{nicefrac}       
\usepackage{enumitem}          
\usepackage{microtype}         
\usepackage{soul}              
\usepackage{graphicx}          
\usepackage{wrapfig}           
\usepackage{tcolorbox}
\usepackage{listings}
\tcbuselibrary{listings,breakable}

\usepackage[dvipsnames]{xcolor}         
\definecolor{bluegray}{rgb}{0.67, 0.9, 0.93}
\definecolor{babypink}{rgb}{0.96, 0.76, 0.76}
\definecolor{lightgreen}{rgb}{0.56, 0.93, 0.56}
\definecolor{mygreen}{rgb}{0.0, 0.5, 0.0}
\definecolor{light_purple}{HTML}{dfccfa}
\definecolor{vermilion}{rgb}{0.776, 0.396, 0.149}     
\definecolor{yellowish}{rgb}{0.933, 0.894, 0.380}      
\definecolor{skyblue}{rgb}{0.435, 0.698, 0.894}        

\newtcblisting{mintedbox}[1][]{
  colback=white,
  colframe=black!70,
  listing only,
  left=5pt,
  breakable,
  boxrule=0.5pt,
  arc=4pt
}

\tcbset{on line,
        boxsep=1pt, left=0pt,right=0pt,top=0pt,bottom=0pt,
        colframe=white,
        colback=light_purple,
        }
\def\paperID{50} 
\def\confName{CVPR}
\def\confYear{2025}

\title{Guiding Diffusion with Deep Geometric Moments: Balancing Fidelity and Variation}

\author{Sangmin Jung\textsuperscript{1}\textsuperscript{$\dagger$} \quad
Utkarsh Nath\textsuperscript{1} \quad
Yezhou Yang\textsuperscript{1} \quad
Giulia Pedrielli\textsuperscript{1} \quad \\
Joydeep Biswas\textsuperscript{2} \quad
Amy Zhang\textsuperscript{2} \quad
Hassan Ghasemzadeh\textsuperscript{1} \quad
Pavan Turaga\textsuperscript{1} \quad \\
\textsuperscript{1}Arizona State University \quad
\textsuperscript{2}The University of Texas at Austin
}

\begin{document}

\twocolumn[{
\renewcommand\twocolumn[1][]{#1}
\maketitle
\includegraphics[width=\textwidth]{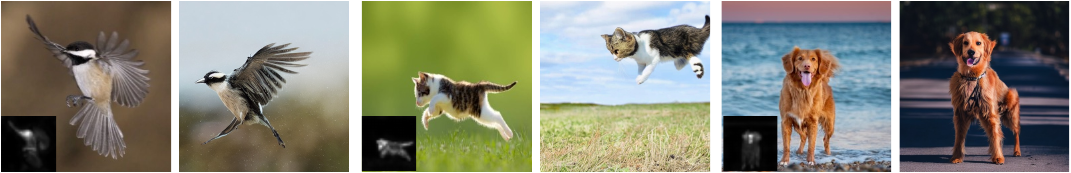}
\captionof{figure}{Deep Geometric Moments (DGM) are used as guidance to capture nuanced subject details from the reference image while preserving the generative diversity of the diffusion model. Each generated result is conditioned on the prompt “a photo of <{\em animal name}>” For each pair, the left image is the reference, and the right image is the generated result. The small image in the bottom-left corner of the reference is the DGM visualization used for guidance.}
\label{fig:intro}
\vspace{2mm}
}]

\renewcommand{\thefootnote}{}
\footnotetext{\textsuperscript{$\dagger$} Corresponding author: \texttt{sjung61@asu.edu}}


\begin{abstract}
Text-to-image generation models have achieved remarkable capabilities in synthesizing images, but often struggle to provide fine-grained control over the output. Existing guidance approaches, such as segmentation maps and depth maps, introduce spatial rigidity that restricts the inherent diversity of diffusion models. In this work, we introduce Deep Geometric Moments (DGM) as a novel form of guidance that encapsulates the subject’s visual features and nuances through a learned geometric prior. DGMs focus specifically on the subject itself compared to DINO or CLIP features, which suffer from overemphasis on global image features or semantics. Unlike ResNets, which are sensitive to pixel-wise perturbations, DGMs rely on robust geometric moments. Our experiments demonstrate that DGM effectively balance control and diversity in diffusion-based image generation, allowing a flexible control mechanism for steering the diffusion process.
\end{abstract}
    
\section{Introduction}

Diffusion models have emerged as a powerful text-to-image (T2I) generation framework, enabling high-quality and diverse synthesis from user input text prompts~\cite{patel2024eclipse, ho2020denoising, patel2024eclipsemain}. Early diffusion methods relied on input gradients from a pre-trained classifier model to guide the generation process. This need was later eliminated by Classifier-free-guidance~\cite{ho2022classifier, Sadat2024NoTN}, which integrated the conditioning mechanism directly into the generation process. Many large-scale image generation models, such as Stable Diffusion~\cite{rombach2022high} and DALL·E~\cite{ramesh2021zero}, advance text-to-image generation using CLIP encoders and cross-attention to align text and image features, achieving generation with much higher fidelity and realism. Despite their impressive generative quality, they often lack fine-grained control over output.

To address this, methods like ControlNet~\cite{zhang2023addingconditionalcontroltexttoimage} and IP-Adapter~\cite{ye2023ip} introduce auxiliary control structures, such as depth maps, Canny edges, segmentation masks, and pose estimations. These methods learn to steer the generation process by applying cross-attention with embeddings of control structures.  However, these methods require dedicated training phases to align these guidance signals with the generation pipeline. Recent works have introduced training-free methods for guided image generation ~\cite{Zampini2025TrainingFreeCG}. These methods do not require retraining but modify the sampling process or leverage external models. Various types of guidance have been explored~\cite{huang2024classdiffusion, nair2023steered, luo2024readout}, including placing bounding boxes to specify regions to generate, providing face IDs (with distinct facial features like nose and lips) to replicate human faces, utilizing CLIP text guidance, and applying style transfer based on CLIP features. 

\begin{figure}[!h]
  \centering
  \includegraphics[width=\columnwidth]{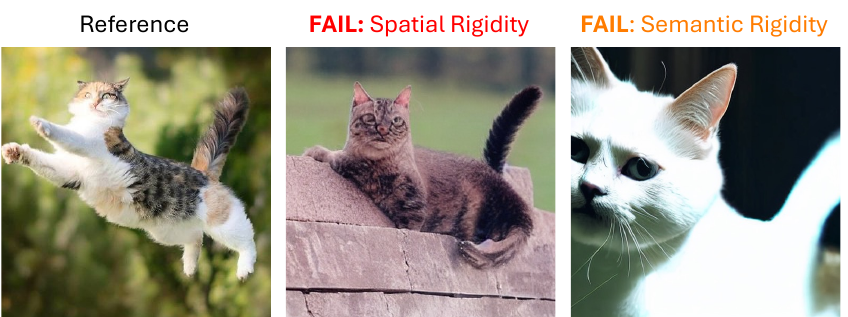}
  \caption{Failure cases of guidance based methods. Spatial rigidity uses segmentation maps as guidance~\cite{bansal2023universal}; Semantic rigidity shows results of CLIP image features as guidance.}
  \label{fig:failure}
\end{figure}

Despite recent advancements, training-based and training-free methods predominantly rely on control signals or global features such as CLIP~\cite{radford2021learning}. As shown in Figure~\ref{fig:failure}, this often leads to generated imagery that either rigidly follow the control signal or firmly adhering to the semantic alignment, failing to preserve the fine-grained visual features of the reference object. To enable a more diverse yet fidelity-preserving generation, it is essential that image generation models not only align with the intended control but also retain creative flexibility beyond rigid structural constraints. 

To this extent, we propose using geometric moments for guidance in a training-free fashion. Geometric moments define various shape characteristics and can be visualized as projections of the image onto chosen basis functions. We propose a guidance algorithm that leverages an auxiliary off-the-shelf pretrained Deep Geometric Moment (DGM) \cite{singh2023improving, nath2024polynomial} model during the image sampling process of a diffusion model. The DGM approach is suited to extract features capable of capturing object textures and other features in a spatially invariant manner allowing for diverse generation while preserving subject identity. Our contribution can be summarized as:

 \begin{itemize}[noitemsep]
\item We introduce the Deep Geometric Moments (DGM) as a unique guidance signal for the diffusion process, enabling effective preservation of the visual details of subjects.
\item We present our pipeline in a training-free setting, eliminating the need for training networks for guided generation by utilizing an off-the-shelf pretrained DGM model.
\item We demonstrate that our generated results maintain the diversity of the diffusion model, unlike other fixed control methods such as segmentation maps or depth maps.
\end{itemize}
\section{Related Works}
\textbf{Conditioned Generation} approaches generally rely on training diffusion models specifically tailored to various conditions using class labels~\cite{dhariwal2021diffusion}. For instance, Ho et al. \cite{ho2022classifier} proposed classifier-free guidance, where diffusion models are trained to blend conditional and unconditional outputs according to provided class labels. Extending the concept further, CLIP Guidance~\cite{nichol2022glidephotorealisticimagegeneration} introduced the use of detailed textual descriptions as prompts, aligning generated images with text embeddings from CLIP. However, their reliance on training separate classifiers limits practical scalability.

\textbf{Guided Generation} methods, in contrast, aim to keep diffusion models fixed and modify only the sampling procedure to achieve targeted outcomes without retraining. Methods like Universal Guidance~\cite{bansal2023universal}, FreeDOM~\cite{yu2023freedom} introduced flexible training-free frameworks that integrate various guidance signals such as CLIP embeddings for style transfer, segmentation maps for spatial control, face ids for replicating human face, demonstrating broader applicability. Further works have been proposed to improve the operation of noise estimation for steering the output ~\cite{song2023loss, he2023manifold}, and analyze an algorithm-agnostic design space of training-free guidance altogether ~\cite{ye2024tfg, Guo2025TrainingFreeGB, Xu2025RethinkingDP, Daras2024ASO, Shen2024UnderstandingAI, Song2024UnravelingTC}.
\section{Method}
\textbf{Deep Geometric Moments (DGM)} \cite{singh2023improving, nath2024polynomial} are a feature representation designed to capture the geometric moment information from images. Geometric moment computes shape descriptors by integrating pixel-wise features over a spatial domain:

\begin{equation}
M_{p,q} = \int\int x^p y^q f(x,y) \,dx\,dy,
\end{equation}

where, \(M_{p,q}\) represents the geometric moment of order \((p, q)\), and \(f(x,y)\) is the function describing the image features at location \((x, y)\). Unlike traditional moment-based descriptors, which rely on predefined basis functions, DGM leverages deep neural networks to learn hierarchical representations that encode geometric attributes of objects. By applying neural networks to extract and refine these moment-based features, DGM produces representations that are robust to variations in scale, rotation, and appearance.

\begin{table*}[t]
    \caption{\textbf{Quantitative comparison of different methods.} We distinguish between training-based and training-free methods. Methods listed below the first dashed line are implemented on top of Universal Guidance. “Prompt Ada.” (Prompt Adaptability) indicates whether a method can incorporate changes in text prompts. CLIP-I measures visual consistency, I-DINO measures diversity, and ChatGPT scores reflect image quality to supplement CLIP-I. For all metrics, higher scores correspond to better performance. However, for I-DINO score nearing 0.5 seem to imply inconsistent output as seen in qualitative results in figure \ref{fig:qualitative-comparison}.}
    \label{tab:quantitative}
    \centering
    \resizebox{\textwidth}{!}{
    \begin{tabular}{l cc c r@{\hspace{0.5em}}l c}
    \toprule
    \textbf{Models} & \textbf{Training-Free} & \textbf{Prompt Ada.} & \textbf{CLIP-I (↑)} & \multicolumn{2}{c}{\textbf{I-DINO (↑)}} & \textbf{ChatGPT (↑)} \\
    \midrule
    \textbf{SD-Img2Img (RGB)} & $\times$ & \checkmark & 0.8601 & 0.2922 & \textcolor{OliveGreen}{(balanced)} & 1.88 \\
    \textbf{IP-Adapter (RGB)} & $\times$ & \checkmark & 0.8947 & 0.1086 & \textcolor{blue}{(stagnant)} & 2.63 \\
    \textbf{ControlNet-Depth (RGB+Depth)} & $\times$ & \checkmark & 0.8606 & 0.1819 & \textcolor{blue}{(stagnant)} & 2.05 \\
    \hdashline
   
    \textbf{DINO} & \checkmark & $\times$ & 0.8735 & 0.1682 & \textcolor{blue}{(stagnant)} & 2.18 \\
    \textbf{ResNet34} & \checkmark & $\times$ & 0.7844 & 0.4847 & \textcolor{red}{(divergent)} & 1.41 \\
        \hdashline

     \textbf{Segmentation Maps} & \checkmark & \checkmark & 0.7431 & 0.5582 & \textcolor{red}{(divergent)} & 0.69 \\
    \textbf{CLIP} & \checkmark & \checkmark & 0.7480 & 0.4787 & \textcolor{red}{(divergent)} & 0.39 \\
    \rowcolor{Apricot!50}\textbf{Ours} & \checkmark & \checkmark & 0.8323 & 0.2754 & \textcolor{OliveGreen}{(balanced)} & 1.85 \\
    \bottomrule
    \end{tabular}
    }
\end{table*}

\textbf{Guidance Sampling with DGM Features}
To steer generation towards specific outputs based on the reference input, we incorporate universal guidance algorithm using features from a pre-trained DGM encoder. Traditional classifier guidance modifies the predicted noise using the gradient of a classifier's log-likelihood:
\begin{equation}
\epsilon_{\theta}(z_t, t) \leftarrow \epsilon_{\theta}(z_t, t) - \sqrt{1 - \alpha_t} \nabla_{z_t} \log p_{\phi}(c|z_t),
\end{equation}
where, \( \epsilon_\theta \) is the guided noise prediction, and \( p_\phi(c | z_t) \) is the probability of class \( c \) predicted by a classifier \( p_\phi \). However, this method has a limitation such that the classifiers are not trained on noisy latent images \( z_t \), and thus they cannot reliably predict class labels at intermediate steps of the diffusion process. We instead compute the classifier signal on the estimated clean image \( \hat{z}_0 \), derived at each step using the DDIM~\cite{song2020denoising} formulation:
\begin{equation}
\hat{z}_0 = \frac{z_t - \sqrt{1 - \alpha_t} \epsilon_\theta(z_t, t)}{\sqrt{\alpha_t}}.
\end{equation}
This step reconstructs a noise-free image estimate \( \hat{z}_0 \) from the current noisy latent \( z_t \), enabling the application of classifiers or feature extractors trained on clean images.
We leverage the clean estimate \( \hat{z}_0 \) to provide meaningful feedback for guidance sampling.
We extend classifier guidance by replacing the classifier with a feature extractor \( f \), and guiding generation using a feature-specific loss function \(l\):
\begin{equation}
\epsilon_{\theta}(z_t, t) \leftarrow \epsilon_{\theta}(z_t, t) + s(t) \cdot \nabla_{z_t}, \ell(c, f(\hat{z}_0))
\end{equation}
where, \( s(t) \) is a scaling factor controlling the strength of guidance, and the loss function for our pipeline is defined as:
\begin{equation}
\ell = \text{MSE}(f(z_{\text{ref}}), f(\hat{z}_0)).
\end{equation}
Here, \( z_{\text{ref}} \) is a reference image with the target subject, and \( f \) extracts visual features from the estimated clean image. This formulation enables guidance based on each feature's distinct characteristics, rather than relying on less fine-grained and less informative raw class labels.
As observed in prior works~\cite{chung2022diffusion, patel2024steering}, a single-step correction is often insufficient to effectively steer the generation process. Therefore, we adopt a repetitive per-step correction strategy following the Universal guidance. Specifically, we re-inject random Gaussian noise \( \epsilon \sim \mathcal{N}(0, I) \) into \( z_{t-1} \) to obtain \( z_t' \) as follows:
\begin{equation}
z_t' = \sqrt{\alpha_t / \alpha_{t-1}} \cdot z_{t-1} + \sqrt{1 - \alpha_t / \alpha_{t-1}} \cdot \epsilon.
\end{equation}

This formulation ensures that \( z_t' \) has the appropriate noise scale corresponding to timestep \( t \), allowing for more controlled guidance by better exploring the distribution at the subsequent step.

\begin{figure*}[t]
  \centering
  \includegraphics[width=\textwidth]{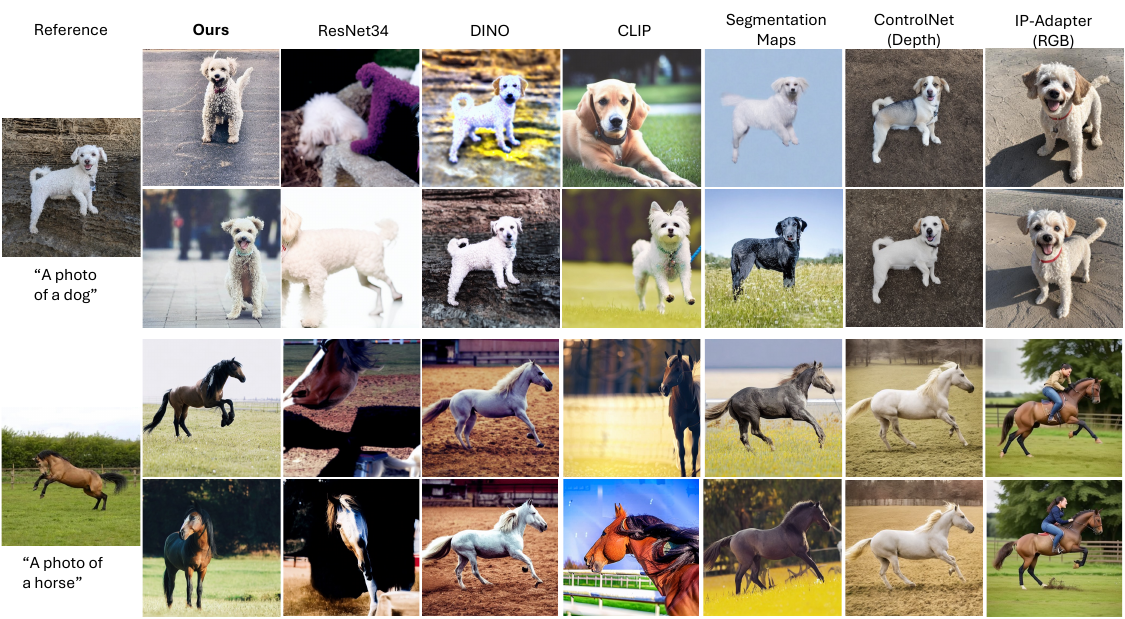}
  \caption{\textbf{Qualitative Comparison of different methods.} The leftmost image in each row is the reference image. Our method, which uses DGM as guidance, is followed by four alternative feature descriptors and two training-based methods. Overall, our DGM-based guidance achieves a balanced trade-off between visual consistency and generative diversity in the diffusion output.}
  \label{fig:qualitative-comparison}
\end{figure*}

\section{Experiments}
\subsection{Experimental Setting} 
\textbf{Datasets and Baselines.}
We curated a unique dataset, \textit{DGMBench}, which focuses on animals with distinct features and nuanced visual details. Our baseline evaluations cover three key aspects: (1) the use of diverse feature descriptors for guidance, (2) the distinction between training-based and training-free methods, and (3) the incorporation of various input modalities beyond RGB images, including segmentation and depth maps. Additional details about the dataset and the experimental setup are provided in the supplement.

\subsection{Quantitative Comparison}
We evaluate the image generation quality based on two main criteria: the ability to preserve fine-grained visual details from the reference image, and the capacity to produce diverse outputs during the diffusion process.
We report the CLIP-I score and GPT-based~\cite{achiam2023gpt} preference scores to measure visual consistency. CLIP-I score is the average cosine similarity between the CLIP image embeddings of the reference and generated images. However, CLIP-I alone may not capture all aspects of visual fidelity. To complement it, we introduce a preference score using the \textit{ChatGPT-4o-latest} model on image inputs. The model ranks the outputs based on pre-defined qualitative rubrics from 0 to 4. Further details on this evaluation setup are provided in the supplementary material.
To quantify variation, we propose I-DINO, which uses the inverse of the average pairwise similarity across DINO\cite{amir2021deep} embedding of the generated images.

\begin{equation}
\begin{aligned}
\operatorname{I ~\text{-}DINO} =
1 - \frac{1}{n} \sum_{i,j}^{n} \operatorname{DINO}(\text{img}_i, \text{img}_j)
\end{aligned}
\end{equation}

Table~\ref{tab:quantitative} summarizes the results across visual fidelity (CLIP-I), variation (I-DINO), and perceived image quality (ChatGPT score). Among the training-free methods, our approach achieves the second-highest CLIP-I score after DINO. However, unlike DINO, which replicates the reference image, our method strikes a better balance at preserving subject identity while enabling diverse outputs. In contrast, the best performing training-based methods are limited in diversity and often generate highly similar output, and sometimes fail to capture the subject's fine features. Our DGM-guided method and SD-Img2Img preserve visual fidelity while also producing more varied generation, demonstrating the effectiveness of DGM in balancing fidelity and variation.

\subsection{Qualitative Comparison}
For the qualitative comparison, we present Figure~\ref{fig:qualitative-comparison}, which showcases the results of our method. Our approach demonstrates that DGM effectively captures the nuanced details and textures of the reference image, whereas naive ResNet~\cite{he2016deep} features are more prone to failure due to their lack of robustness against pixel-wise variations. DINO features tend to replicate the input reference image almost entirely, but lack flexibility because of their strong patch-wise global matching behavior. CLIP maintains semantic alignment; however, preserving fine-grained visual details remains challenging. The segmentation maps and the ControlNet depth results show that both methods can maintain the object's location according to their design purpose, but occasionally fail by overly constraining the generated results to a designated area. Overall, IP-Adapter performs well, though it often renders visual details with a specific stylistic bias and sometimes introduces undesired elements, such as generating a human figure on a horse, as shown in the figure. In general, subject identity is captured more accurately than with simple guidance-based methods. Additional examples are provided in the supplemental materials.
\section{Conclusion}
In this work, we introduce a novel guidance method, Deep Geometric Moments (DGM), for guiding text-to-image generation. We demonstrate that utilizing a feature vector from the DGM model effectively transfers the visual details of the subject. Using this feature, we successfully maintain visual consistency with the reference image while preserving the diffusion model’s output diversity. Our approach is more effective than the feature guidance from ResNet, which often fails due to its sensitivity to pixel-level perturbations. We also conduct an extensive comparison with various feature descriptors, such as DINO and CLIP, revealing their limitations, which include adhering too closely to global image representation or overly emphasizing semantic consistency.
In addition, although training-based methods achieve higher fidelity to their inputs, they significantly reduce the generative diversity inherent to diffusion models. Through extensive quantitative and qualitative evaluations, our results demonstrate that our approach effectively balances subject identity preservation and output diversity~\cite{gandikota2025distilling} without requiring additional training.

\section*{Acknowledgements}
This paper was supported by NSF grant 2402650 and in part by NSF grant 2323086. We thank the Research Computing (RC) at Arizona State University (ASU) and \href{https://www.cr8dl.ai}{cr8dl.ai} for their generous support in providing computing resources. We would also like to express our gratitude to Suren Jayasuriya for insightful guidance that informed the development of this project.


{
    \small
    \bibliographystyle{ieeenat_fullname}
    \bibliography{main}
}

\newpage
\clearpage
\setcounter{page}{1}
\maketitlesupplementary

\section{Preliminaries for Diffusion models}
The forward process in diffusion models can be formulated as:

$$ q(x_t | x_{t-1}) = \mathcal{N}(x_t; \sqrt{\alpha_t} x_{t-1}, (1 - \alpha_t) I) $$
where $x_t$ denotes predicted latent at timestep t, $\alpha_t$ is the pre-defined variance schedule and $I$ is the identity matrix.
Then our $x_t$ equation can be also written equivalently as:
$$x_t = x_0 \sqrt{\bar{\alpha}_t} + \epsilon \sqrt{1 - \bar{\alpha}_t}, \quad \epsilon \sim \mathcal{N}(0, I)$$
Then our neural network with trainable parameters $\theta$, and our reverse step samples:
$$p_\theta(x_{t-1} | x_t) := \mathcal{N} \left( x_{t-1}; \mu_\theta(x_t, t), \Sigma_\theta(x_t, t) \right)
$$
which can be obtained by minimizing the variational lower bound of the negative log likelihood (ELBO) of the distribution. 

\section{Dataset Details}
For evaluating the task of replicating the visual information of the subject in the input reference image, we curated our dataset, DGMBench. DGMBench focuses on subjects with fine details. The dataset was primarily created by utilizing a subset of the Dreambench dataset~\cite{ruiz2023dreambooth}, including dogs and cats, while excluding general non-fine-grained subjects such as vases and bowls. Additionally, we included a variety of animals with diverse appearances, such as birds and fish, while avoiding uniformly looking animals like elephants and lions. The dataset consists of 30 subjects in total.

\section{Experiment Details}
\textbf{Rationale of selecting Universal Guidance as backbone}
Through empirical experiments, we observed that the Universal Guidance mechanism outperforms other guidance methods and inverse problem solvers in diffusion, with expense of some additional computational overhead. ~\cite{zhang2024improving, chung2022diffusion}\\
\textbf{Explanation of feature descriptors}
The original DGM model was trained with an architectural structure (3, 4, 6, 3 intermediate layers), replicating the ResNet34 design, for the same goal of image classification. We selected ResNet34 as a comparison to evaluate the performance of its learned features and to validate whether ResNet-trained features are sufficiently robust to serve as effective inputs for the guidance loss.
DINO image features, trained in large-scale self-supervised settings, are widely used across various computer vision tasks but due to its adherence to capture the whole image representation, not utilized for purposes such as guiding the generation. In contrast, CLIP is primarily for its use with its text features, which have been employed in methods such as GLIDE. However, in our experiments, we focus on CLIP's image features to assess their effectiveness for our tasks. Segmentation maps are also included as part of the baseline Universal Guidance approach, and we evaluate their use for showing the diversity in generation.
Based on empirical evaluation, the guidance scales and recurrence steps for each method were selected as shown in Table~\ref{tab:guidance_rec_steps}.
\begin{table}[htbp]
\centering
\resizebox{\columnwidth}{!}{%
\begin{tabular}{lcccc}
\toprule
\textbf{Method} & \textbf{Guidance Scale} & \textbf{Recurrence Steps} & \textbf{Feature Size} & \textbf{Loss Function} \\
\midrule
Ours   & 10000 & 10 & [1, 256]    & MSE     \\
ResNet & 2000  & 10 & [1, 512]          & MSE     \\
DINO   & 1000  & 10 & [197, 384]  & CosSim  \\
CLIP   & 10    & 10 & [1, 768]    & CosSim  \\
Seg    & 400   & 10 & img\_size   & CE      \\
\bottomrule
\end{tabular}
}
\caption{Guidance scale and number of recurrence steps used for different methods.}
\label{tab:guidance_rec_steps}
\end{table}

To ensure a consistent evaluation environment, we adopt Stable Diffusion v1.5 as the common backbone for all baseline methods, including ControlNet-Depth, SD-Img2Img, and IP-Adapter. In addition, the number of sampling steps is fixed at 500 across all methods to unify the experiment conditions.

\section{Prompt for ChatGPT Evaulation}
\lstset{
  basicstyle=\ttfamily\small,          
  backgroundcolor=\color{gray!10},      
  keywordstyle=\color{blue},            
  commentstyle=\color{green!50!black},  
  stringstyle=\color{red},              
  numbers=left,                         
  numberstyle=\tiny\color{gray},        
  frame=single,                         
  breaklines=true,                      
  captionpos=b                          
}


\begin{lstlisting}
[Task]
You are a human evaluator assessing visual similarity between a reference image and a comparison image. Your focus is on low-level visual consistency - this includes color patterns, texture fidelity, and fine-grained inner geometric details (such as fur, stripes, surface patterns).
Ignore factors like the subject's pose, overall shape, or semantic realism.

[Evaluation]
Please assign a score from 0 to 4 based on how well the comparison image preserves the visual essence of the reference image. Use the following rubric:

[Evaluation Criteria]
Score from 0-4 based on:
0	Completely inconsistent =-Major differences in color, texture, and detail. The comparison image does not preserve any recognizable fine details from the reference.
1	Weak consistency - Some minor resemblance in visual traits, but largely different. Texture or color patterns are either distorted or missing.
2	Moderate consistency - There are identifiable shared traits (like fur direction or partial texture), but notable mismatches in color tones or local structures.
3	Strong consistency - Most key visual attributes like texture patterns, color palettes, and local details are preserved, with only slight differences.
4	Excellent consistency - Nearly identical in terms of texture, color, and fine-grained visual cues. Differences, if any, are negligible and hardly noticeable.

[Output]
Using the above rubric, compare the following two images and return a single score (0-4), along with a one-sentence justification of the score.
\end{lstlisting}

\section{Limitations and Failure Cases}
\begin{figure}[!h]
  \centering
  \includegraphics[width=\columnwidth]{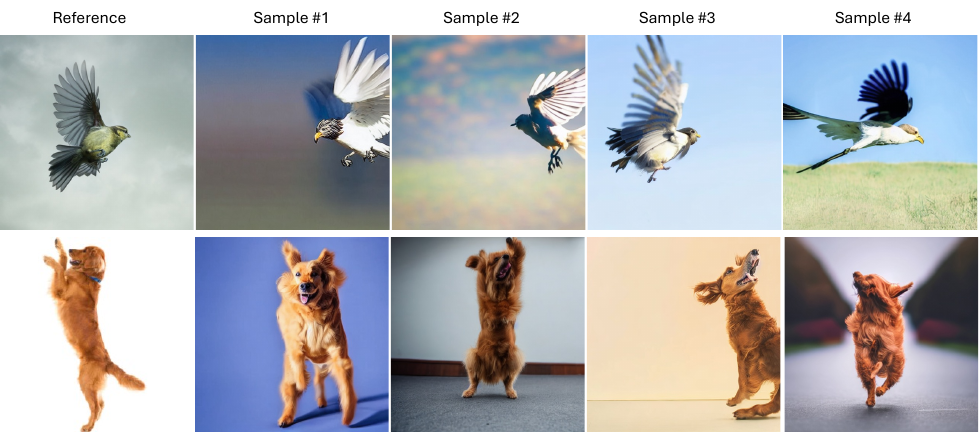}
  \caption{Failure examples of our method.}
  \label{fig:failure_ours}
\end{figure}

The performance of training-free guidance is highly dependent on the choice of feature descriptors and their combination with the guidance parameters. The baseline image generator, Stable Diffusion, occasionally produces unwanted artifacts when the guidance strength is too high or when the guidance does not properly align with the target distribution of the diffusion model, making careful hyperparameter selection essential. Subtle mismatches between the guidance signal and the diffusion process can further result in unnatural images or artifacts, as observed in the CLIP- and ResNet-guided results.\\
As with other models, our model fails when the subject in the reference image has dynamic shapes or poses. As shown in Figure~\ref{fig:failure_ours}, when the original dog is standing on its hind legs or the bird is flying in a dynamic position, such semantic difficulties lead to degraded results. Additionally, if the subject has intricate features of various details and textures, the results fail, presumably due to the learned geometric prior being insufficient.

\section{Additional Qualitative Results}
More extensive qualitative comparison is shown in following Figures~\ref{fig:add_qual_0}--\ref{fig:add_qual_3}. For optimal clarity, the figures are best viewed in color prints.

\section{Future Works}
Based on our findings, we hypothesize that DGM can be leveraged as a prior to enhance the performance of existing personalized, subject-driven text-to-image (T2I) models, a direction we leave for future work.

\clearpage
\begin{figure*}[p]
    \centering
    \includegraphics[width=\textwidth,height=\textheight,keepaspectratio]{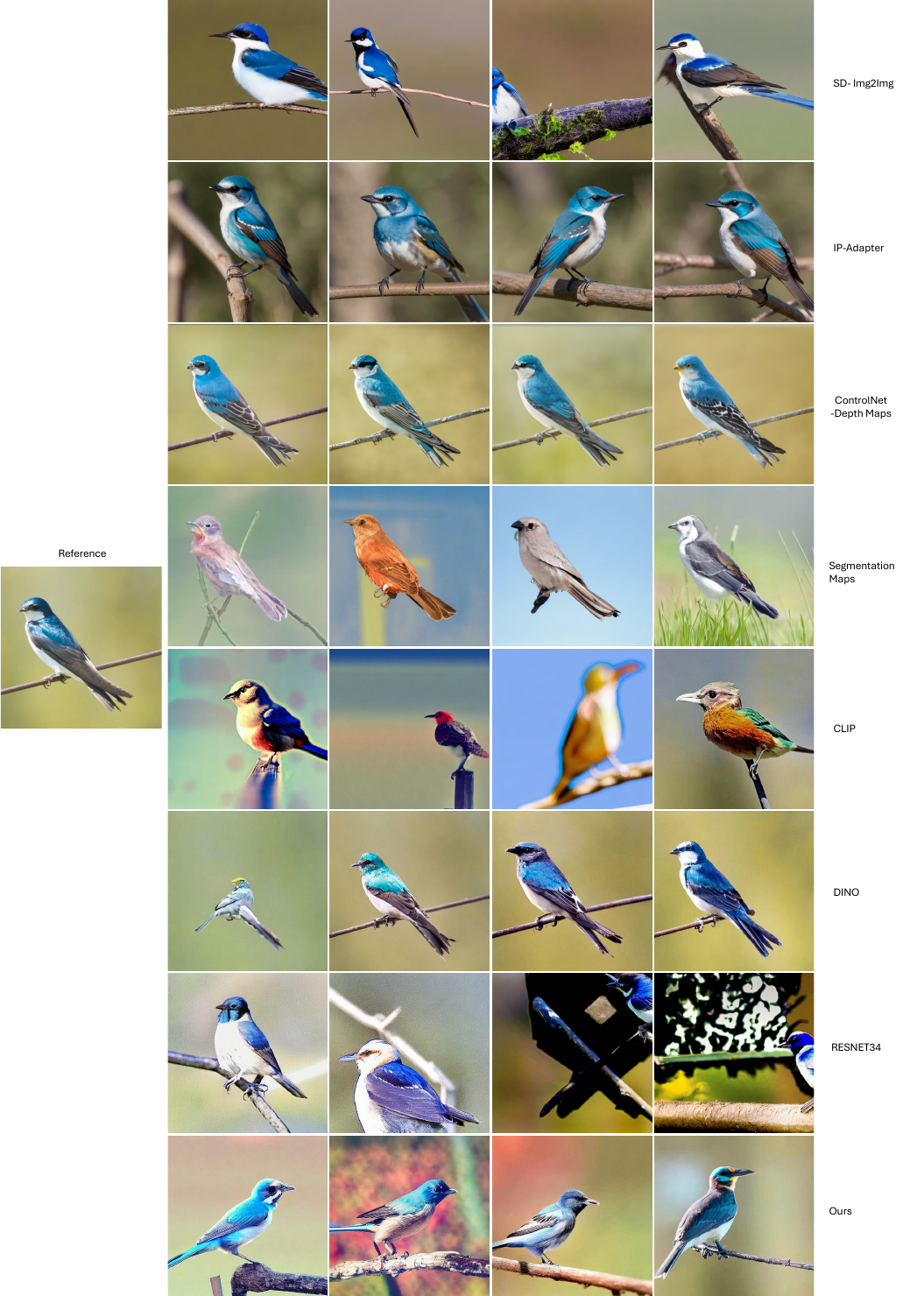}
    \caption{\textbf{Additional qualitative comparison results} for the prompt: "A photo of a bird". None of the results are cherry-picked.}
    \label{fig:add_qual_0}
\end{figure*}

\begin{figure*}[p]
    \centering
    \includegraphics[width=\textwidth,height=\textheight,keepaspectratio]{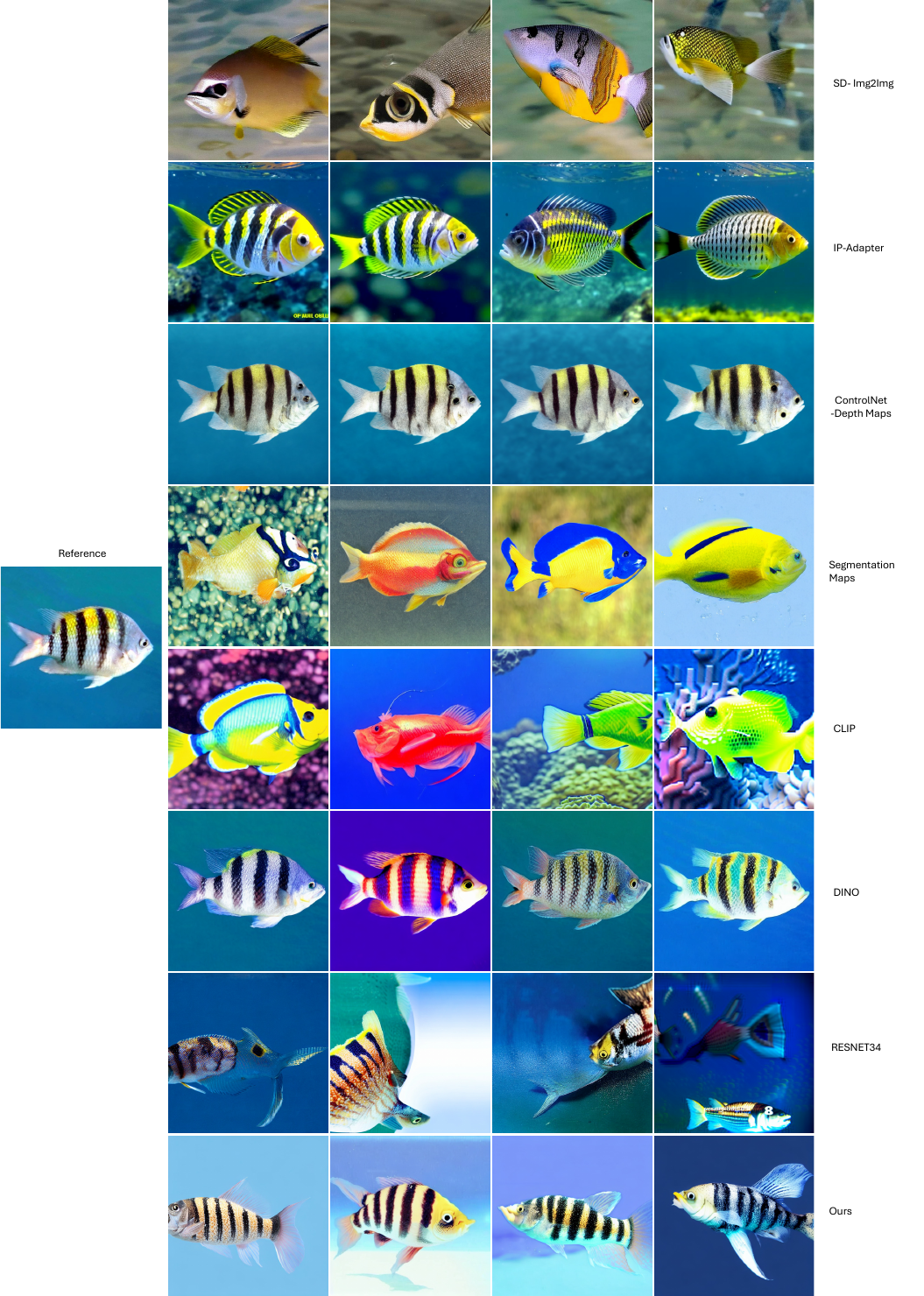}
     \caption{\textbf{Additional qualitative comparison results} for the prompt: "A photo of a fish". None of the results are cherry-picked.}
    \label{fig:add_qual_1}
\end{figure*}

\begin{figure*}[p]
    \centering
    \includegraphics[width=\textwidth,height=\textheight,keepaspectratio]{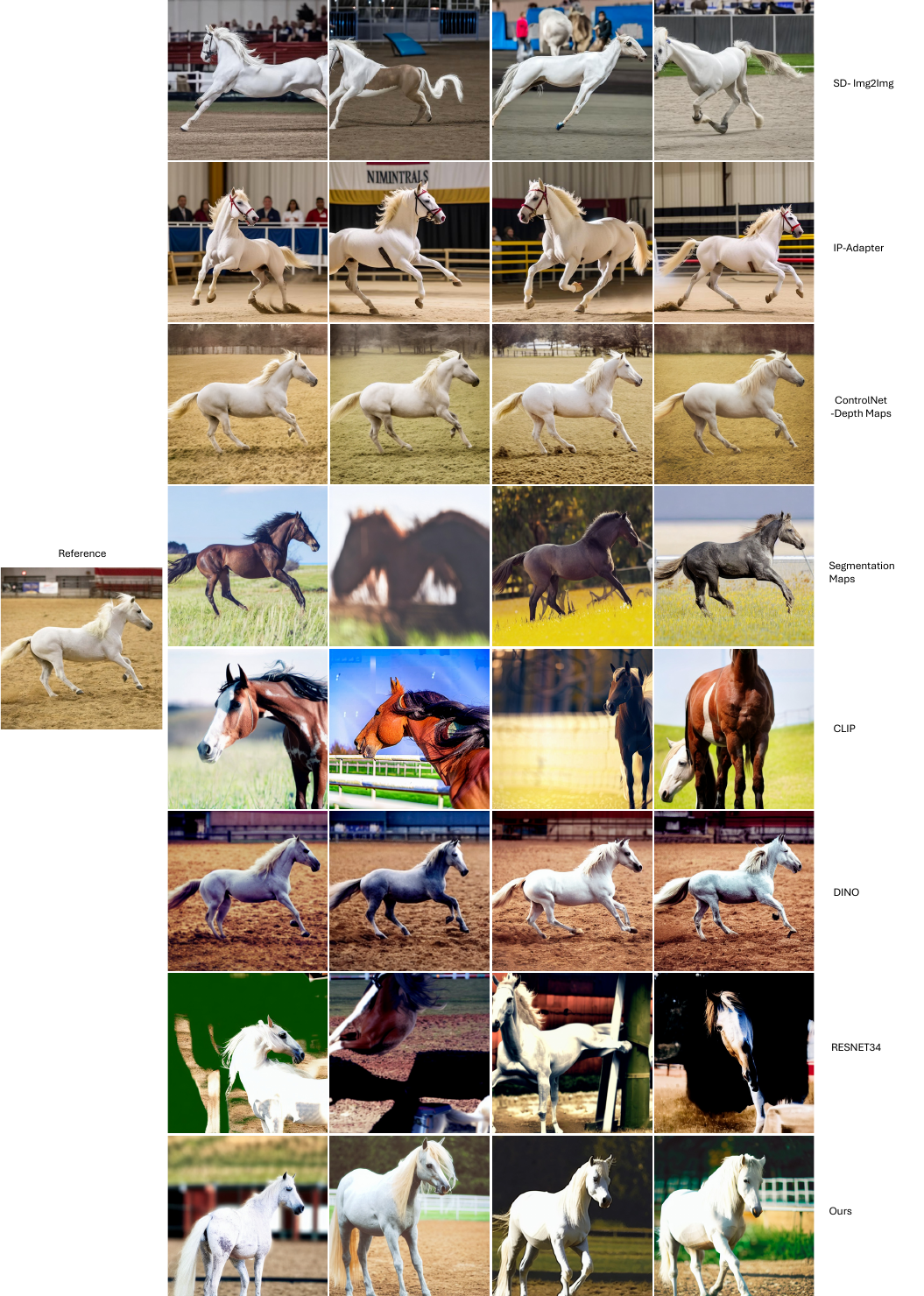}
     \caption{\textbf{Additional qualitative comparison results} for the prompt: "A photo of a horse". None of the results are cherry-picked.}
    \label{fig:add_qual_2}
\end{figure*}

\begin{figure*}[p]
    \centering
    \includegraphics[width=\textwidth,height=\textheight,keepaspectratio]{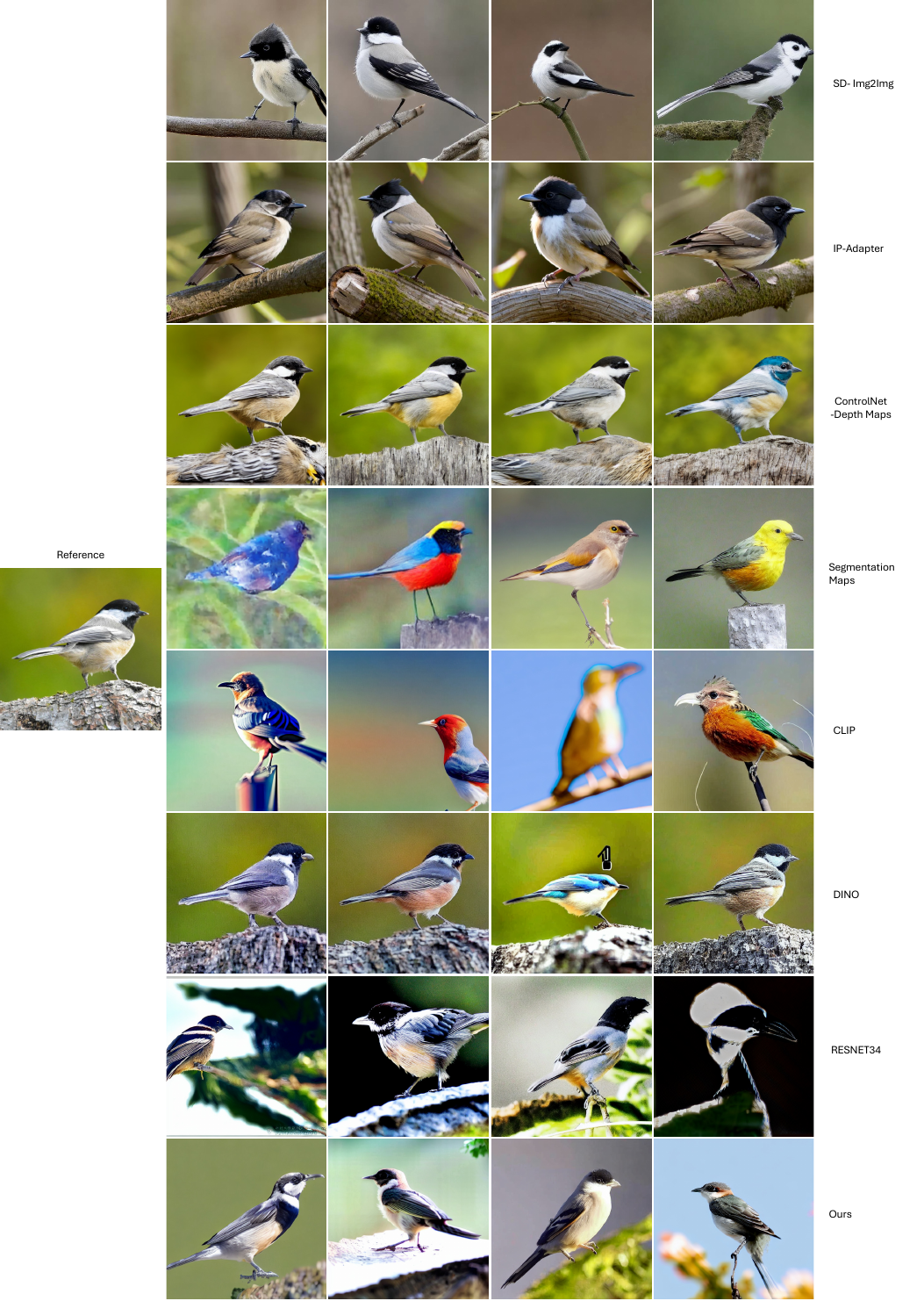}
    \caption{\textbf{Additional qualitative comparison results} for the prompt: "A photo of a horse". None of the results are cherry-picked.}
    \label{fig:add_qual_3}
\end{figure*}


\end{document}